\documentclass{article}

\usepackage{PRIMEarxiv}

\usepackage[utf8]{inputenc} 
\usepackage[T1]{fontenc}    
\usepackage{hyperref}       
\usepackage{url}            
\usepackage{booktabs}       
\usepackage{amsfonts}       
\usepackage{nicefrac}       
\usepackage{microtype}      
\usepackage{lipsum}
\usepackage{fancyhdr}       
\usepackage{graphicx}       
\graphicspath{{media/}}     
\usepackage{xcolor}
\usepackage{algorithm2e}
\usepackage{amsmath} 
\usepackage{afterpage}
\usepackage{authblk}
\usepackage[symbol*]{footmisc}
\usepackage{tocloft}
\usepackage{multirow}

\title{Enabling High-Sparsity Foundational Llama Models with Efficient Pretraining and Deployment
}

\author{
    \begin{tabular}{c | c | c | c}
        Abhinav Agarwalla*\textsuperscript{1} \; Abhay Gupta*\textsuperscript{2} \; Alexandre Marques*\textsuperscript{1} \; Shubhra Pandit*\textsuperscript{1}
    \end{tabular}
    \begin{tabular}{c | c | c | c | c}
         Michael Goin\textsuperscript{1} \; Eldar Kurtic\textsuperscript{1} \;   Kevin Leong\textsuperscript{2} \; Tuan Nguyen\textsuperscript{1} \;   Mahmoud Salem\textsuperscript{2} 
    \end{tabular}
    \begin{tabular}{c | c | c}
         Dan Alistarh\textsuperscript{1,3} \; Sean Lie\textsuperscript{2} \; Mark Kurtz\textsuperscript{1} 
    \end{tabular}
    \begin{tabular}{c}
         
    \end{tabular}
    \begin{tabular}{c | c | c}
         \textbf{\textsuperscript{1}Neural Magic} \;   \textbf{\textsuperscript{2}Cerebras Systems} \;   \textbf{\textsuperscript{3}IST Austria}
    \end{tabular}
}

\begin{document}
\maketitle

\begin{abstract}
Large language models (LLMs) have revolutionized Natural Language Processing (NLP), but their size creates computational bottlenecks. We introduce a novel approach to create accurate, sparse foundational versions of performant LLMs that achieve full accuracy recovery for fine-tuning tasks at up to 70\% sparsity. We achieve this for the LLaMA-2 7B model by combining the SparseGPT one-shot pruning method and sparse pretraining of those models on a subset of the SlimPajama dataset mixed with a Python subset of The Stack dataset. We exhibit \emph{training acceleration} due to sparsity on Cerebras CS-3 chips that closely matches theoretical scaling. 
In addition, we establish \emph{inference acceleration} of up to 3x on CPUs by utilizing Neural Magic's DeepSparse engine and 1.7x on GPUs through Neural Magic's nm-vllm engine. The above gains are realized via sparsity alone, thus enabling further gains through additional use of quantization. 
Specifically, we show a total speedup on CPUs for sparse-quantized LLaMA models of up to \textit{8.6x}. We demonstrate these results across diverse, challenging tasks, including chat, instruction following, code generation, arithmetic reasoning, and summarization to prove their generality. This work paves the way for rapidly creating smaller and faster LLMs without sacrificing accuracy.

\vspace{0.5cm}

The resulting models, code, and documentation are open sourced at the following links to promote the reproducibility and expansion of our results:
\begin{itemize}
    \item \href{https://docs.neuralmagic.com/llms/models/sparse-foundational-llama-2}{Code and Documentation}
    \item \href{https://huggingface.co/collections/neuralmagic/sparse-foundational-llama-2-models-65f48cec6396309f02e74d21}{HuggingFace Models Collection}
\end{itemize}
\end{abstract}

\vfill

\footnote{
    Equal contribution. Corresponding authors: abhinav@neuralmagic.com, abhay@cerebras.net, alexandre@neuralmagic.com, shubhra@neuralmagic.com
}

\begin{figure}[htbp]
    \centering
    \makebox[\textwidth][c]{\includegraphics[width=1.1\textwidth]{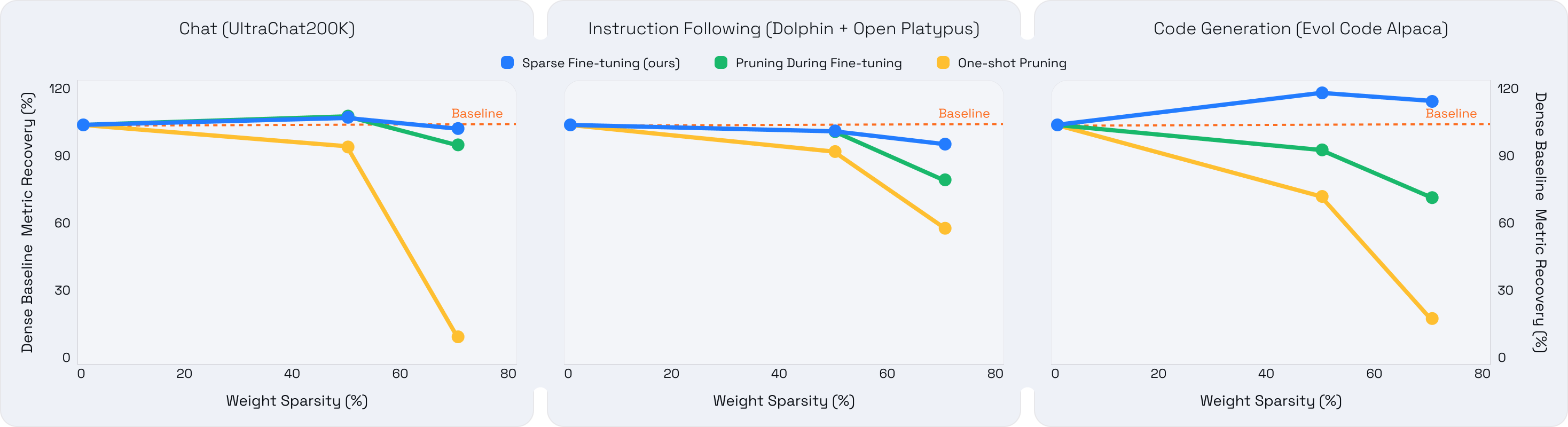}}
    \caption{Sparse Llama-2 7B performance summaries showing sparsity vs. recovery level across chat, instruction following, and code generation.}
    \label{fig:sparse_deploy_vs_recovery}
\end{figure}

\section{Introduction}

Large language models (LLMs) have transformed the field of Natural Language Processing (NLP) \cite{vaswani2017attention}. Their ability to generate text, translate languages, and answer questions in a near-human way has opened up unprecedented applications. However, their massive size creates a significant bottleneck. The computation cost of training and running these models is very high, has significant energy impact, and limits their accessibility \cite{zhu2023climate, luccioni2022estimating, mosaic2022llmcosts}. Compression techniques such as quantization have successfully reduced model size and improved inference speed \cite{lin2020towards, dettmers2022llm, frantar2022gptq}. However,  quantization past 4 bits per parameter while preserving accuracy is proving to be a limit that is hard to cross for full recoverability compared to high-quality baseline models \cite{yao2022zeroquant,dettmers2023spqr,chee2023quip}.

Weight pruning, that is, setting a fraction of the model parameters to zero, is another promising approach to reach higher compression in the context of Deep Neural Networks (DNNs) \cite{hoefler2021sparsity,kurtic2023ziplm,voita-etal-2019-analyzing,kurtic2022optimal}. 
Specifically, sparsity reduces the model's storage footprint and can enable faster inference and training through reduced computation and memory movement. However, existing pruning methods often struggle to maintain high accuracy, especially at high sparsity levels and complex tasks \cite{kurtic2023sparse}. These  accuracy limitations reduce their potential for creating genuinely efficient and generally usable sparse LLMs. 
To our knowledge, no techniques currently exist for accurately pruning foundational models to non-trivial sparsities while preserving their abilities on downstream tasks.   

To address this challenge, we consider a new approach that combines accurate pruning and fine-tuning of a foundational model, which we illustrate on the Llama-2 7B architecture \cite{touvron2023llama}. Specifically, we investigate the following steps:
\begin{itemize}
    
\item \textbf{Sparse Pretraining}: We introduce a new approach to creating sparse LLMs that achieves high accuracy for fine-tuned models at up to 70\% sparsity. Our approach expands on top of the popular SparseGPT \cite{frantar2023sparsegpt} post-training pruning algorithm with further pretraining of the sparse models on subsets of the popular SlimPajama~\cite{cerebras2023slimpajama} and The Stack~\cite{kocetkov2022thestack} datasets. We demonstrate that this combination enables higher recovery levels after fine-tuning than the current state-of-the-art (SOTA) pruning during fine-tuning with distillation \cite{kurtic2023sparse}. The combination was particularly impactful for more challenging tasks such as chat, code generation, and instruction following.

\item \textbf{Practical Speedups for Training and Inference}: We demonstrate practical speedups for the entire LLM lifecycle, including training and deployment. We show close to ideal speedups for sparse training utilizing Cerebras's CS-3 AI accelerator \cite{cerebras2024cs3}. Further, we use Neural Magic's software stack to deploy SOTA sparse performance on CPUs and GPUs through DeepSparse \cite{deepsparse} and nm-vllm \cite{nmvllm}, respectively.

\item \textbf{Compounding Gains with Quantization}: We show that our sparse foundational models can be further quantized while maintaining accuracy, enabling compounding performance gains. These results highlight the synergistic potential of combining sparsity with quantization.

\end{itemize}

Figure \ref{fig:sparse_deploy_vs_recovery} provides a high-level visualization of our results. Our sparse LLMs and efficient compute platforms offer dramatic speedups while preserving accuracy and creating a valuable stepping stone toward widely accessible and highly performant LLMs.

\section{Methodology}

\subsection{Sparse Pretraining}

\begin{algorithm}[H]
\SetAlgoLined
\KwIn{Dataset $\mathcal{D}$, sparse model $\theta$, mask $\mathbf{M}$, learning rate $\eta$, num\_steps}
\KwOut{Converged sparse model $\theta$}

\For{step in range(num\_steps)}{
    (x, y) = random\_sample($\mathcal{D}$) \tcc*[r]{Mini-batch sample}
    $\hat{y} = \text{forward}(\mathbf{x}, \theta)$ \tcc*[r]{Calculate prediction}
     $\mathcal{L} = \text{loss}(\hat{y}, y)$  \tcc*[r]{Calculate loss}
    $\nabla_{\theta} \mathcal{L} = \text{backward}(\mathcal{L})$  \tcc*[r]{Calculate gradients}
    $\nabla_{\theta} = \nabla_{\theta} \odot \mathbf{M}$ \tcc*[r]{Zero out gradients with mask}
    $\theta = \theta - \eta \nabla_{\theta}$ \tcc*[r]{Weight update}
    $\theta = \theta \odot \mathbf{M}$ \tcc*[r]{Zero out weights with mask} 
}
\caption{Sparse Pretraining Algorithm to ensure sparsity masks are held constant.} 
\label{alg:sparse-pretraining}
\end{algorithm}

In a post-training setting, we employed SparseGPT \cite{frantar2023sparsegpt} to introduce sparsity into the Llama-2 7b model, targeting a 50\% sparsity level with a uniform per-layer sparsity profile. To obtain the 70\% sparsity level, we adopt an iterative approach: we first train the 50\% sparse model until convergence, and only then prune it additionally to obtain the 70\% target. After the pruning step we freeze the pruned weights and enforce sparsity mask during training to preserve sparsity. For completeness, we illustrate this process in Algorithm~\ref{alg:sparse-pretraining}. In addition to the uniform sparsity profile we experimented with the more recent Outlier Weighed Layerwise (OWL) sparsity profiles~\cite{yin2024outlier}. As shown in Appendix Section~\ref{app:ablations}, based on our limited ablation study, the uniform sparsity profile performed slightly better in the pretraining phase. 

We selected the SlimPajama dataset \cite{cerebras2023slimpajama} mixed with Python subset of The Stack~\cite{kocetkov2022thestack} for sparse pretraining due to their filtered nature, ensuring a high percentage of quality tokens. For more details on the pretraining dataset mixture, please refer to the Appendix Section~\ref{app:ablations}.
For the 50\% sparse model, we utilized 45 billion tokens of pretraining data, while an additional 100 billion tokens were used for the 70\% model. This represents approximately 2\% to 8\% of the original 2 trillion tokens used to train the base Llama-2 model.

\subsection{Sparse Fine-Tuning}

Building on the SquareHead distillation approach \cite{kurtic2023sparse}, we integrated it with our pretrained sparse models to enable sparse fine-tuning with per-layer distillation. This combination proved crucial in achieving high accuracy on complex tasks at higher sparsities. Furthermore, at lower levels of sparsity in combination with distillation, we note that we can recover higher than the baseline accuracy for simpler tasks, following along the lines of previous research that has shown that moderate sparsity can act as a regularizer \cite{jin2022prunings}.

\textbf{Fine-Tuning Experiment Types:}
We evaluated the following four sparse fine-tuning approaches:
\begin{itemize}
    \item \textit{Dense Fine-Tuning with One-Shot Pruning:} Fine-tuning a dense model followed by one-shot pruning on the fine-tuning dataset.
    \item \textit{Pruning During Fine-Tuning:} Additional sparse fine-tuning was applied to the models from the first approach.
    \item \textit{Sparse Fine-Tuning from One-Shot Pruned Models:} Pruning the pretrained models followed by sparse fine-tuning onto the target dataset.
    \item \textit{Sparse Fine-Tuning from Sparse Pretrained Models:} Sparse fine-tuning on the fine-tuning dataset, starting from our sparse pretrained models.
\end{itemize}

\textbf{Task-Specific Recovery Trends:}
We observed distinct recovery patterns across different task categories, suggesting a correlation between task complexity and effective fine-tuning strategies:
\begin{itemize}
    \item \textit{Limited Context Tasks:} Fine-tuning with pruning often achieves full recovery for limited-context tasks (e.g., arithmetic reasoning, summarization). This suggests the fine-tuning dataset largely contains the information needed for model adaptation.
    \item \textit{Large Context Tasks:} Recovery is significantly more challenging for large context tasks (e.g., chat, code generation, instruction following) using standard pruning during fine-tuning, indicating a greater reliance on broader knowledge from the pretraining dataset.
\end{itemize}

\textbf{Advantages of Sparse Pretraining:} Our experiments highlighted the advantages of our sparse pretraining and subsequent sparse fine-tuning methodology compared to the standard "pruning during fine-tuning" approach:
\begin{itemize}
    
\item \textit{Higher Recovery at High Sparsities}: Particularly for complex tasks with large context windows (chat, code generation, instruction following), our sparse pretraining approach consistently demonstrated superior accuracy recovery at up to 70\% sparsity levels.

\item \textit{Simplified Hyperparameter Search}: Sparse pretraining creates a more robust foundation, often reducing the need for extensive hyperparameter tuning typical of pruning during fine-tuning.

\item \textit{Reduced Computation}: Utilizing sparse pretrained models often allows for a single fine-tuning run to achieve convergence. This result contrasts with the "pruning during fine-tuning" pathway, which typically involves converging a dense model, followed by pruning and subsequent additional fine-tuning.

\end{itemize}

\subsection{Sparse Pretraining Acceleration on Cerebras CS-3}

\begin{figure}[htbp]
    \centering
    \includegraphics[width=0.75\textwidth]{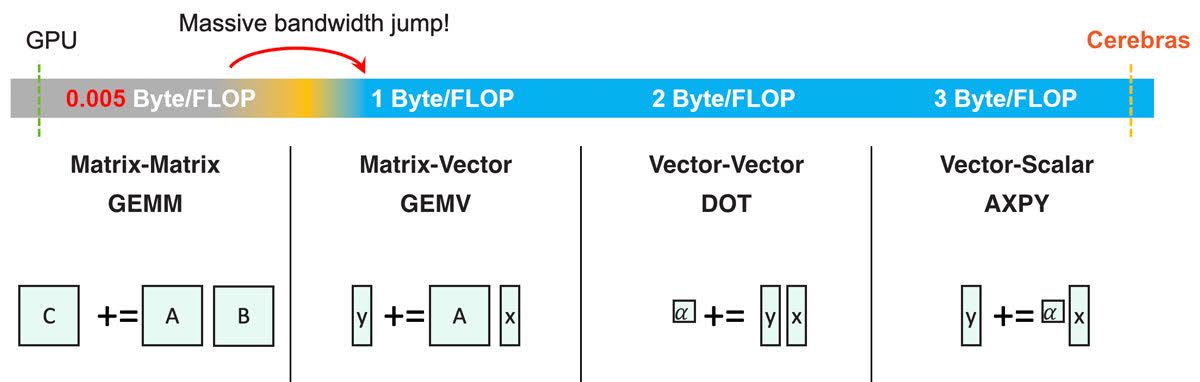}
    \caption{Sparse GEMM, essential for sparse LLM training, has particularly high bandwidth demands. The Cerebras CS-3 architecture outperforms traditional systems with its unique on-chip memory to address this bottleneck.} 
    \label{fig:cs3_memory_bandwidth} 
\end{figure}

The Cerebras CS-3 AI accelerator \cite{cerebras2024cs3} is uniquely designed to accelerate sparse training with native support for unstructured sparsity. Its foundation is a fully programmable processor supporting general-purpose and specialized tensor instructions, enabling flexibility and performance for diverse workloads. Sparse general matrix multiply (GEMM) operations, at the heart of LLM training, benefit from the CS-3's unique on-chip memory architecture. This architecture delivers the high memory bandwidth essential for sparse operations, overcoming the limitations of traditional GPUs reliant on off-chip DRAM, as shown in Figure \ref{fig:cs3_memory_bandwidth}.

Fine-grained dataflow execution further empowers the CS-3 for sparse acceleration. In this paradigm, data itself triggers computation. With zero values filtered out, only non-zero data is processed, leading to power savings and accelerated performance. This dataflow mechanism seamlessly supports unstructured sparsity, as found through our sparse pretraining methodology.

\begin{figure}[htbp]
    \centering
    \includegraphics[width=0.6\textwidth]{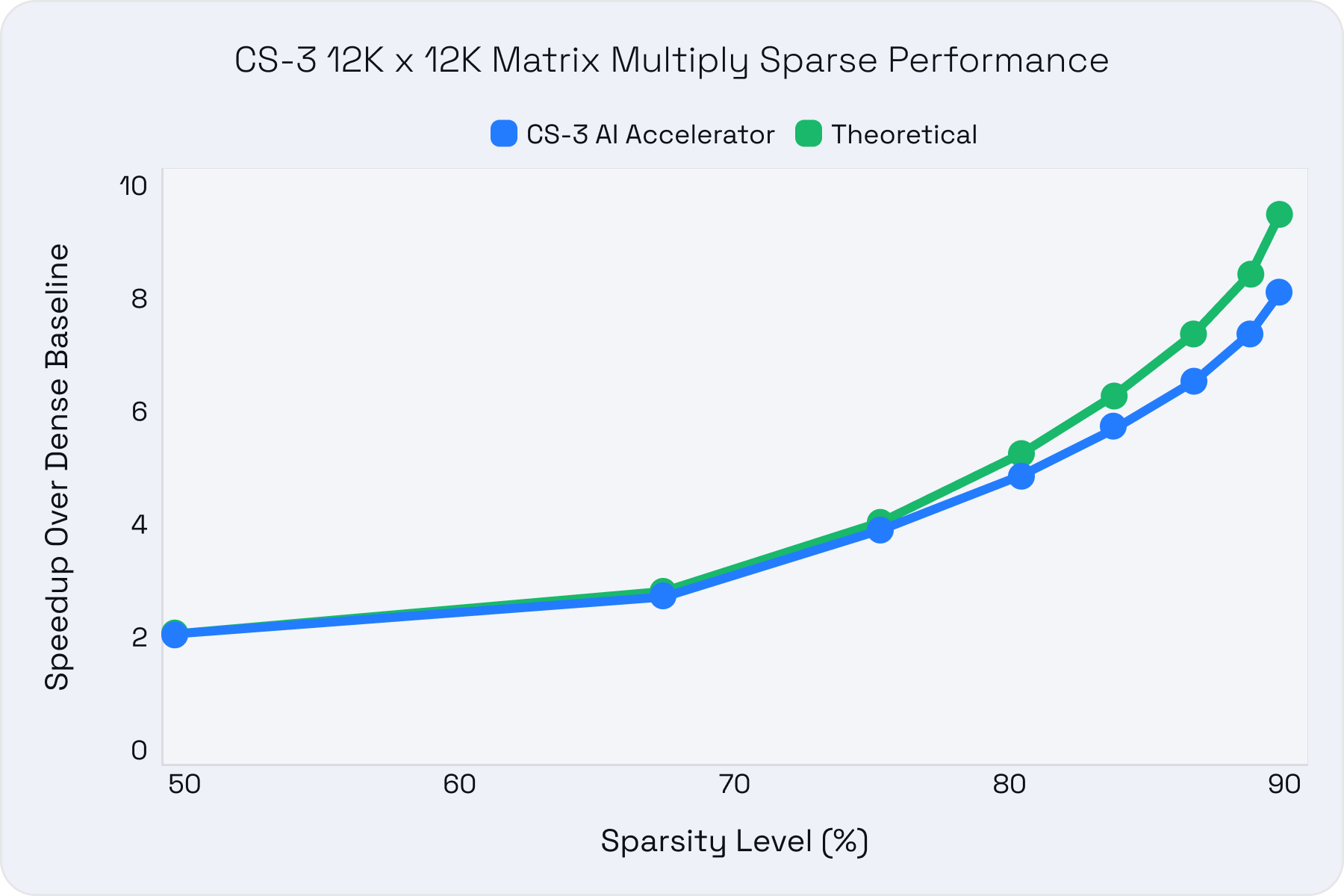}
    \caption{Compute performance for a 12K x 12K Matrix Multiply on Cerebras CS-3 AI accelerator comparing theoretical vs. realized performance in relation to sparsity.}
    \label{fig:sparse_scaling} 
\end{figure}

The result is a significant reduction in theoretical FLOPs and a tangible increase in LLM training performance on our systems. The seamless integration with PyTorch ensures accessibility, allowing developers to leverage the hardware's capabilities without extensive code modifications. Our implementation closely mirrors the expected theoretical scaling, as Figure \ref{fig:sparse_scaling} demonstrates, showcasing an example matrix multiplication from the Llama-2 architecture. For a more in-depth technical exploration please refer to Cerebras architecture paper \cite{lie2022cerebras}. 

\vspace{4cm}

\subsection{Sparse Inference Acceleration}

Achieving performant sparse inference on CPUs and GPUs poses a unique challenge, particularly with unstructured sparsity. Below, we summarize our approaches and optimizations for each architecture.

\subsubsection{CPU Implementation}

Efficient sparse inference on CPUs necessitates specialized techniques to address the challenges of unstructured sparsity. Neural Magic's DeepSparse Engine \cite{deepsparse} employs bitmask expansion to optimize sparse weight representation and achieve inference speedups. This technique compresses memory usage and enhances decoding performance.

In bitmask expansion, we operate on a per-SIMD-block basis, where SIMD stands for "Single Instruction Multiple Data", leveraging the parallel processing capabilities of modern CPUs. Popular SIMD instruction sets include AVX-512 and VNNI, which DeepSparse actively utilizes for implementation. Non-zero values within each block are stored densely, and a bitmask indicates the sparsity pattern. This format enables efficient storage and retrieval of sparse weights, fully utilizing the CPU's SIMD resources.

\textbf{Example:} 

\begin{minipage}{\textwidth}
\begin{verbatim}
Consider the following 2x8 matrix of 16-bit integers:
+----+-----+----+----+----+----+----+----+
| 0  | A2  | 0  | A4 | A5 | 0  | 0  | A8 |
+----+-----+----+----+----+----+----+----+
| B1 | B2  | 0  | 0  | 0  | B6 | B7 | B8 |
+----+-----+----+----+----+----+----+----+

Corresponds to the following register layout for matrix-vector multiplication on VNNI:
[ 0, B1, A2, B2, 0, 0, A4, 0, A5, 0, 0, B6, 0, B7, A8, B8 ]

Resulting in the following sparse representation utilizing bitmasking:
values: [ B1, A2, B2, A4, A5, B6, B7, A8, B8 ]
bitmask: 0b0111001010010111
\end{verbatim}
\end{minipage}
\newline

\textbf{Benefits:} Bitmask expansion in DeepSparse reduces memory footprint, even at moderate sparsity levels (e.g., 50\%), which is crucial for large language models. Additionally, it facilitates efficient memory access patterns, maximizing CPU performance. 

\begin{figure}[htbp]
    \centering
    \includegraphics[width=0.6\textwidth]{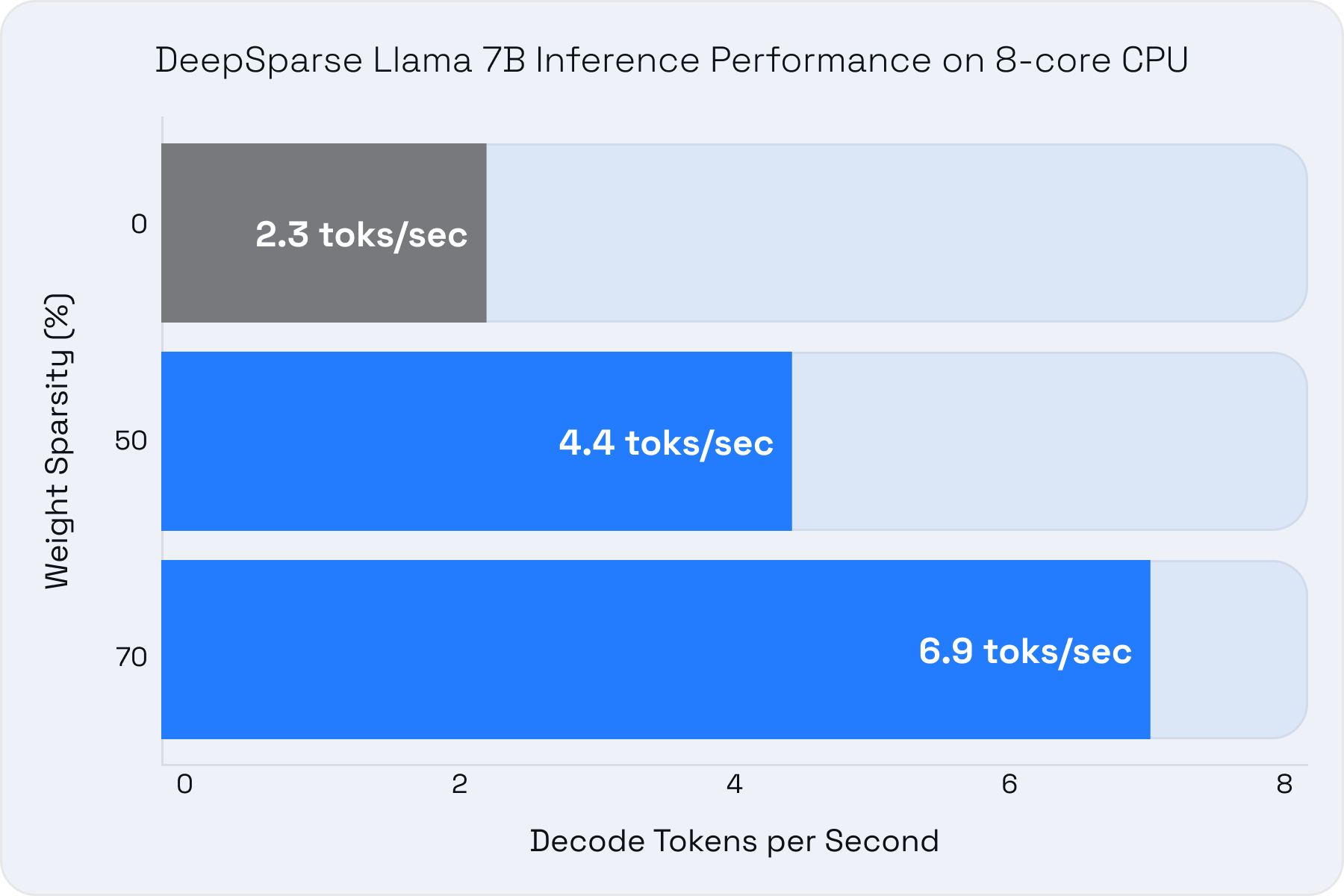}
    \caption{Decode inference performance for varying sparsity levels with Llama 7B utilizing DeepSparse on an 8-core AMD CPU (AWS c7a.4xlarge).}
    \label{fig:deepsparse-sparse-performance} 
\end{figure}

\subsubsection{GPU Implementation}

Achieving high-performance sparse inference on GPUs requires exploiting their inherent parallelism and specialized matrix multiplication hardware. Neural Magic's nm-vllm \cite{nmvllm} extends the bitmask expansion concept to matrices, optimizing for tensor core utilization. Like the CPU approach, sparse weight matrices are compressed using dense values and bitmasks. However, the focus is on efficiently using GPU register files and tensor cores as the target.

Compressed sparse matrices reside in GPU global memory. They are subdivided into smaller submatrices tailored to fit into register files. 
Dense non-zero values within each submatrix are stored with an accompanying sparsity-encoding bitmask. During inference, compressed submatrices are transferred from global memory to register files. The bitmask guides efficient decompression of the dense values, reconstructing the original sparse submatrix within the register files. This process effectively leverages GPU parallelism and bitwise operations.

Once decompressed, the tensor cores directly process the sparse submatrices from the register files. These specialized hardware units excel at matrix multiplication and accumulation, enabling compute performance equivalent to dense matrix operations while significantly reducing global memory bandwidth demands.

\begin{figure}[htbp]
    \centering
    \includegraphics[width=0.6\textwidth]{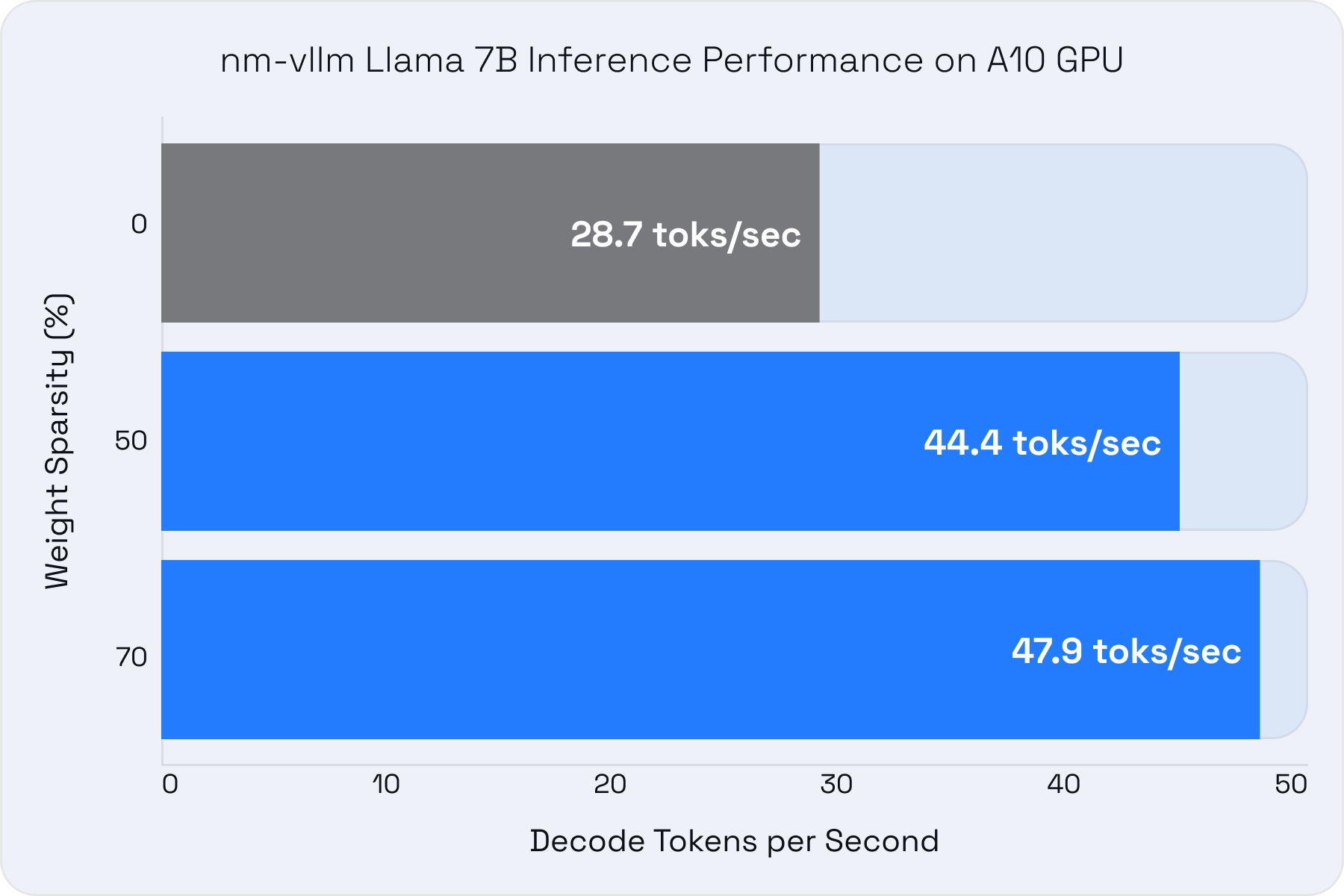}
    \caption{Decode inference performance for varying sparsity levels with Llama 7B utilizing nm-vllm on an A10 GPU (AWS g5.xlarge).}
    \label{fig:deepsparse-sparse-performance} 
\end{figure}

\section{Experimental Validation}

\subsection{Sparse Pretraining}

\begin{table}[htbp]
\centering
\caption{Pretrained Llama-2 7B evaluation metrics compared to varying sparsity levels and approaches. The best results at each sparsity level are highlighted in bold.}
\label{tab:pretraining_recovery}
\includegraphics[width=0.8\textwidth]{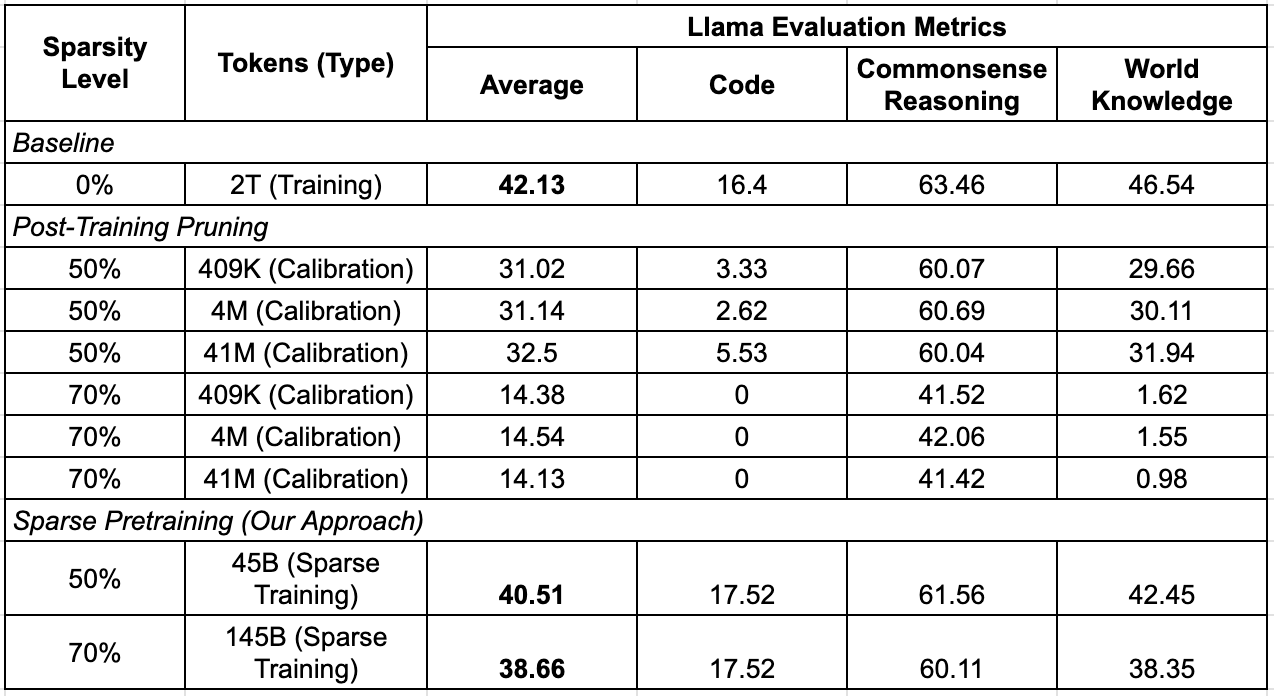}
\end{table}

\textbf{Experimental Setup:}
As noted earlier, our sparse pretraining experiments employed SparseGPT with uniform sparsity profiles. The number of calibration and training tokens was varied to test the effectiveness of further pretraining after pruning. For simplicity, we settled on 1K calibration samples at 4096 sequence length for creating the bases for the sparse pretrained models. 

All pretraining experiments were performed on an 8x Cerebras CS-3 Wafer-Scale Cluster, leveraging standard PyTorch and native data parallel support for distributed training.

\textbf{Sparse Llama-2 7B Performance:}
Our results demonstrate the effectiveness of sparse pretraining in enabling accuracy recovery. Table \ref{tab:pretraining_recovery} compares the SparseGPT post-training pruning-only approach to our sparse pretraining methodology. Post-training pruning alone plateaus around 1000 calibration samples and becomes unstable at 70\% sparsity. In contrast, our sparse pretraining approach achieves significantly higher accuracy recovery:
\begin{itemize}
    
\item \textit{50\% Sparsity}: We achieve 96.1\% recovery of Llama Evaluation metrics with sparse pretraining. This surpasses post-training techniques alone by 19 percentage points, demonstrating the effectiveness of our approach in creating robust, sparse models.

\item \textit{70\% Sparsity}: Our sparse pretraining methodology enables a notable 91.8\% recovery of Llama Evaluation metrics. This represents a significant 57.3 percentage points higher recovery over post-training techniques, highlighting the power of sparse pretraining for challenging high-sparsity regimes.

\end{itemize}

\subsection{Sparse Fine-Tuning for Limited Context Tasks}

\begin{table}[htbp]
\centering
\caption{Fine-tuned Llama-2 7B evaluation metrics across limited context tasks (arithmetic reasoning and summarization) compared to varying sparsity levels and approaches. The best results at each sparsity level are highlighted in bold.}
\label{tab:limited_context_recovery}
\includegraphics[width=0.8\textwidth]{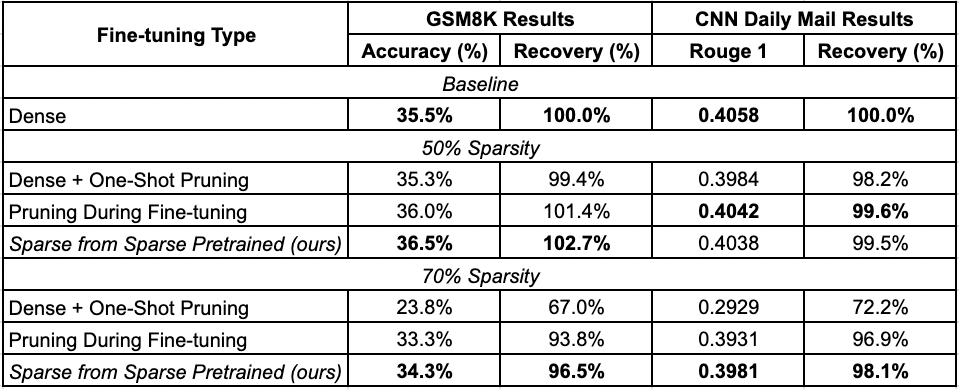}
\end{table}

\textbf{Experimental Setup:} 
To assess the effectiveness of different sparse fine-tuning approaches on limited context tasks, we conducted ablations on GSM8K \cite{gsm8k} (arithmetic reasoning) and CNN Daily Mail \cite{nallapati2016abstractive} (summarization). These datasets were selected to represent task categories where pruning during fine-tuning often yields successful accuracy recovery. The resulting models were evaluated according to their specific tasks. In this case, GSM8K was evaluated based on the zero-shot accuracy on the test set. CNN Daily Mail was evaluated on the validation set utilizing a range of Rouge scores \cite{ng2015better}, including Rouge 1, Rouge 2, and Rouge L. All Rouge scores followed the same trend, and we report Rouge 1 in the table for brevity.

All experiments and evaluations were performed on single-node systems with 8x A100 80GB GPUs, leveraging standard PyTorch, Transformers, and the built-in Fully Sharded Data Parallel (FSDP) support for distributed training.

\textbf{Sparse Llama-2 7B Performance:}
Our results, as detailed in Table \ref{tab:limited_context_recovery}, demonstrate that sparse pretrained models achieve comparable or superior performance to the current SOTA for pruning during fine-tuning, and they excel at high sparsity levels:
\begin{itemize}
    \item \textbf{Robust Recovery:} Sparse pretrained models consistently achieved close to full recovery at both 50\% and 70\% sparsity across tasks. This matches or surpasses the accuracy of models created through pruning during fine-tuning.
    \item \textbf{Simplified Workflow:} Sparse pretraining yields highly accurate sparse models directly from fine-tuning without requiring iterative pruning and fine-tuning cycles. This reduces computational overhead and simplifies the process of creating efficient models.
\end{itemize}

\subsection{Sparse Fine-Tuning for Large Context Tasks}

\begin{table}[htbp]
\centering
\caption{Fine-tuned Llama 2 7B evaluation metrics across large context tasks (chat, instruction following, and code generation) compared to varying sparsity levels and approaches. The best results at each sparsity level are highlighted in bold.}
\label{tab:large_context_recovery}
\includegraphics[width=\textwidth]{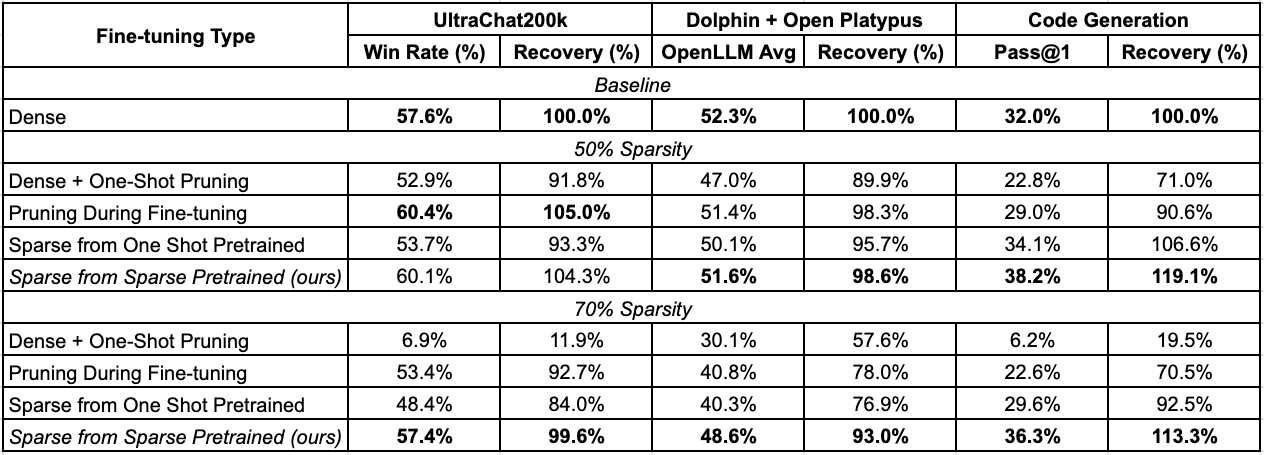}
\end{table}

\textbf{Experimental Setup:}
We conducted ablations on datasets representing large context tasks: UltraChat200k \cite{ding2023enhancing,tunstall2023zephyr} (chat), Dolphin \cite{hartford2023dolphin} + Open Platypus \cite{lee2024platypus} (instruction following), and Evol Code Alpaca \cite{luo2023wizardcoder} (code generation). These tasks were chosen to assess the effectiveness of different sparse fine-tuning techniques when recovery is more challenging. Following both the open-source and research communities' standards, we evaluate chat utilizing AlapacaEval \cite{alpaca_eval} (Llama-2 70B chat evaluator for ease of replication), instruction tuning with OpenLLM benchmarks \cite{open-llm-leaderboard}, and code generation utilizing HumanEval \cite{chen2021codex}. 

All experiments and evaluations were performed on single-node systems with 8x A100 80GB GPUs, leveraging standard PyTorch, Transformers, and the built-in Fully Sharded Data Parallel (FSDP) support for distributed training.

\textbf{Sparse Llama-2 7B Performance:}
Our results demonstrate the significant advantage of sparse pretrained models for large context tasks, especially at high sparsity levels:
\begin{itemize}
    \item \textbf{Superior Recovery:} Sparse pretrained models consistently outperformed all other techniques. At both 50\% and 70\% sparsity, we achieved full recovery on average, with only instruction tuning falling below our threshold at 70\%. This represents a substantial improvement of 21.6 percentage points compared to baseline "pruning during fine-tuning" approaches.
    \item \textbf{Robustness at High Sparsity:} While "pruning during fine-tuning" exhibited significant accuracy degradation at 70\% sparsity, the sparse pretraining methodology proved remarkably resilient.
\end{itemize}

Table \ref{tab:large_context_recovery} provides a detailed breakdown of accuracy metrics across compression levels and techniques, further validating these findings. The higher recovery we see for code generation tasks was thoroughly explored and is attributed to the higher performance of the sparse pretrained models on coding tasks.

\subsection{Sparse Quantized Inference Performance}

\textbf{Adding Post-Training Quantization} To further compress our models, we applied post-training quantization following our sparse fine-tuning workflow. We employed a combination of SmoothQuant \cite{xiao2023smoothquant} and GPTQ \cite{frantar2022gptq} algorithms alongside layer skipping on a model-by-model basis for the top 5 layers (based on kurtosis measurements) to minimize accuracy impact. We emphasized INT8 weight and activation quantization to optimize compatibility with the Neural Magic DeepSparse engine and maximize inference performance gains. However, our results were similar in performance to dense quantization, indicating further quantization schemes would also work.

\textbf{Impact of Quantization}
Notably, our quantization methodology resulted in negligible accuracy degradation across tasks (see the appendix for comprehensive results). The INT8 format for weights and activations is crucial for achieving maximal speedups with the DeepSparse engine, both for time-to-first token and time-per-output token.

\textbf{Performance Improvements}

\begin{figure}[htbp]
    \centering
    \includegraphics[width=\textwidth]{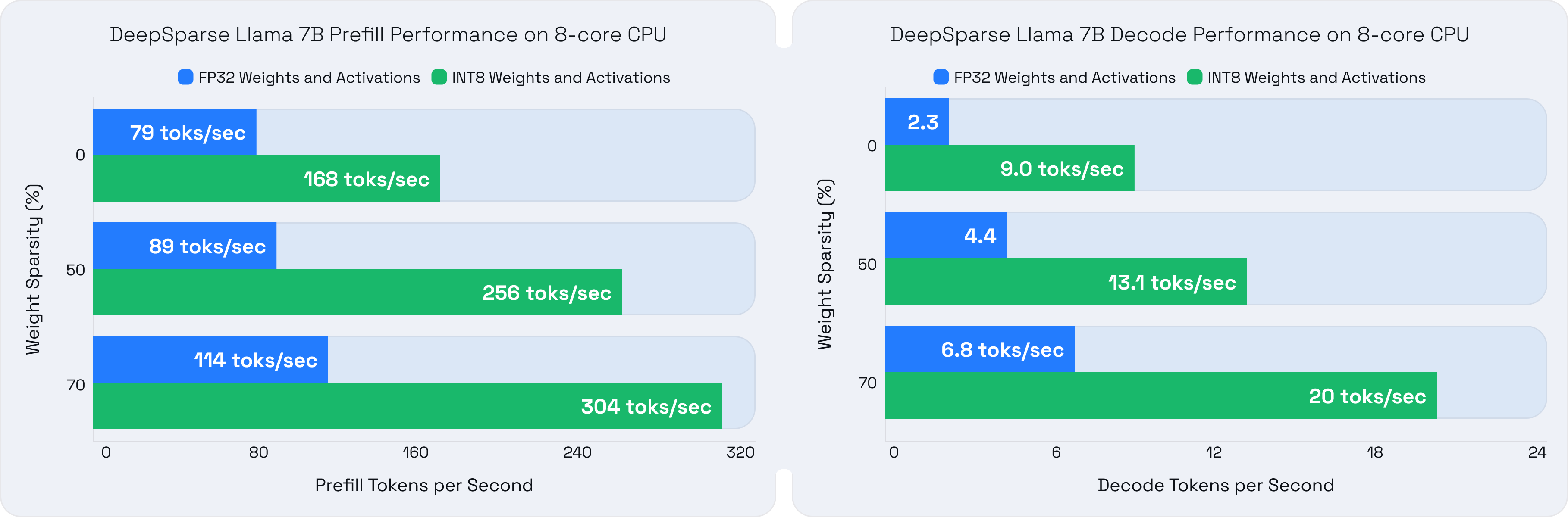}
    \caption{Inference performance for both prefill and decode at varying sparsity and quantization levels with Llama 7B utilizing DeepSparse on an 8-core AMD CPU (AWS c7a.4xlarge).}
    \label{fig:deepsparse-sparse-quantized-performance} 
\end{figure}

Figure \ref{fig:deepsparse-sparse-quantized-performance} visually demonstrates the significant performance gains achieved through our combination of sparsity and quantization. Compared to baseline FP32 models, we observed the following improvements on CPUs:
\begin{itemize}
    \item \textbf{Prefill Performance Increase: } The reduction in compute through INT8 kernels and sparsity decreased time-to-first token by 3.86x for a standard 512 token prefill target.
    \item \textbf{Decode Performance Increase: } Reduced memory size through quantization and sparsity enabled an increase of 8.6x in decode tokens per second.
\end{itemize}

\section{Conclusion and Future Work}

In this paper, we have presented a novel approach for creating sparse foundational LLMs. Our sparse pretraining methodology, pretraining acceleration through Cerebras's CS-3 system, and efficient sparse inference techniques for CPUs and GPUs through Neural Magic's software stack enable significant and holistic performance gains. Notably, we achieve exceptional accuracy recovery even at high sparsity levels (up to 70\%), surpassing traditional pruning during fine-tuning and offering a path toward smaller, faster, and more accessible LLMs.

Our work opens exciting avenues for future research.  These include scaling studies with larger models, adaptations to novel LLM architectures, refined quantization techniques (such as INT4), and investigations into the optimal composition and size of pretraining datasets.  Ultimately, our contributions aim to advance the SOTA of sparse LLM development,  enabling the creation of highly efficient models that maintain accuracy, leading to wider accessibility and the expansion of LLM applications.

\section*{Acknowledgments}

We would like to extend our sincere gratitude to several dedicated teams and individuals who made this research possible:

\begin{itemize}
    \item \textbf{Neural Magic:} Thank you to the entire Neural Magic team, especially those not mentioned in direct authorship working on the DeepSparse engine and supporting software, for their invaluable expertise and commitment to sparse computation.
    \item \textbf{Cerebras Systems: } Thank you to the team at Cerebras Systems, especially those not mentioned in direct authorship working on their cutting-edge CS-3 system supporting unstructured sparsity.
    \item \textbf{IST Austria: } Thank you to the members of Dan Alistarh's lab at IST Austria for their insights and contributions to this project.
\end{itemize}

We deeply appreciate the collective efforts of these teams and individuals in advancing the frontiers of sparse LLM research.

\vfill

\bibliographystyle{unsrt}  
\bibliography{references}  

\begin{thebibliography}{10}

\bibitem{vaswani2017attention}
Ashish Vaswani, Noam Shazeer, Niki Parmar, Jakob Uszkoreit, Llion Jones, Aidan~N Gomez, {\L}ukasz Kaiser, and Illia Polosukhin.
\newblock Attention is all you need.
\newblock In {\em Conference on Neural Information Processing Systems (NeurIPS)}, 2017.

\bibitem{zhu2023climate}
Hongyin Zhu and Prayag Tiwari.
\newblock Climate change from large language models, 2023.

\bibitem{luccioni2022estimating}
Alexandra~Sasha Luccioni, Sylvain Viguier, and Anne-Laure Ligozat.
\newblock Estimating the carbon footprint of bloom, a 176b parameter language model, 2022.

\bibitem{mosaic2022llmcosts}
Abhi Venigalla and Linden Li.
\newblock Mosaic llms: Gpt-3 quality for <\$500k, 2022.

\bibitem{lin2020towards}
Ye~Lin, Yanyang Li, Tengbo Liu, Tong Xiao, Tongran Liu, and Jingbo Zhu.
\newblock Towards fully 8-bit integer inference for the transformer model.
\newblock {\em arXiv preprint arXiv:2009.08034}, 2020.

\bibitem{dettmers2022llm}
Tim Dettmers, Mike Lewis, Younes Belkada, and Luke Zettlemoyer.
\newblock {LLM}.int8(): 8-bit matrix multiplication for transformers at scale.
\newblock {\em Advances in Neural Information Processing Systems 35: Annual Conference on Neural Information Processing Systems 2022, NeurIPS 2022}, 2022.

\bibitem{frantar2022gptq}
Elias Frantar, Saleh Ashkboos, Torsten Hoefler, and Dan Alistarh.
\newblock Gptq: Accurate post-training quantization for generative pre-trained transformers.
\newblock {\em arXiv preprint arXiv:2210.17323}, 2022.

\bibitem{yao2022zeroquant}
Zhewei Yao, Reza~Yazdani Aminabadi, Minjia Zhang, Xiaoxia Wu, Conglong Li, and Yuxiong He.
\newblock Zeroquant: Efficient and affordable post-training quantization for large-scale transformers.
\newblock {\em arXiv preprint arXiv:2206.01861}, 2022.

\bibitem{dettmers2023spqr}
Tim Dettmers, Ruslan Svirschevski, Vage Egiazarian, Denis Kuznedelev, Elias Frantar, Saleh Ashkboos, Alexander Borzunov, Torsten Hoefler, and Dan Alistarh.
\newblock Spqr: A sparse-quantized representation for near-lossless llm weight compression.
\newblock {\em arXiv preprint arXiv:2306.03078}, 2023.

\bibitem{chee2023quip}
Jerry Chee, Yaohui Cai, Volodymyr Kuleshov, and Christopher De~Sa.
\newblock Quip: 2-bit quantization of large language models with guarantees.
\newblock {\em arXiv preprint arXiv:2307.13304}, 2023.

\bibitem{hoefler2021sparsity}
Torsten Hoefler, Dan Alistarh, Tal Ben-Nun, Nikoli Dryden, and Alexandra Peste.
\newblock Sparsity in deep learning: Pruning and growth for efficient inference and training in neural networks.
\newblock {\em arXiv preprint arXiv:2102.00554}, 2021.

\bibitem{kurtic2023ziplm}
Eldar Kurtic, Elias Frantar, and Dan Alistarh.
\newblock Ziplm: Hardware-aware structured pruning of language models.
\newblock {\em arXiv preprint arXiv:2302.04089}, 2023.

\bibitem{voita-etal-2019-analyzing}
Elena Voita, David Talbot, Fedor Moiseev, Rico Sennrich, and Ivan Titov.
\newblock Analyzing multi-head self-attention: Specialized heads do the heavy lifting, the rest can be pruned.
\newblock In {\em Proceedings of the 57th Annual Meeting of the Association for Computational Linguistics}, pages 5797--5808, Florence, Italy, July 2019. Association for Computational Linguistics.

\bibitem{kurtic2022optimal}
Eldar Kurtic, Daniel Campos, Tuan Nguyen, Elias Frantar, Mark Kurtz, Benjamin Fineran, Michael Goin, and Dan Alistarh.
\newblock The {Optimal BERT Surgeon}: Scalable and accurate second-order pruning for large language models.
\newblock {\em arXiv preprint arXiv:2203.07259}, 2022.

\bibitem{kurtic2023sparse}
Eldar Kurtic, Denis Kuznedelev, Elias Frantar, Michael Goin, and Dan Alistarh.
\newblock Sparse fine-tuning for inference acceleration of large language models, 2023.

\bibitem{touvron2023llama}
Hugo Touvron, Louis Martin, Kevin Stone, Peter Albert, Amjad Almahairi, Yasmine Babaei, Nikolay Bashlykov, Soumya Batra, Prajjwal Bhargava, Shruti Bhosale, Dan Bikel, Lukas Blecher, Cristian~Canton Ferrer, Moya Chen, Guillem Cucurull, David Esiobu, Jude Fernandes, Jeremy Fu, Wenyin Fu, Brian Fuller, Cynthia Gao, Vedanuj Goswami, Naman Goyal, Anthony Hartshorn, Saghar Hosseini, Rui Hou, Hakan Inan, Marcin Kardas, Viktor Kerkez, Madian Khabsa, Isabel Kloumann, Artem Korenev, Punit~Singh Koura, Marie-Anne Lachaux, Thibaut Lavril, Jenya Lee, Diana Liskovich, Yinghai Lu, Yuning Mao, Xavier Martinet, Todor Mihaylov, Pushkar Mishra, Igor Molybog, Yixin Nie, Andrew Poulton, Jeremy Reizenstein, Rashi Rungta, Kalyan Saladi, Alan Schelten, Ruan Silva, Eric~Michael Smith, Ranjan Subramanian, Xiaoqing~Ellen Tan, Binh Tang, Ross Taylor, Adina Williams, Jian~Xiang Kuan, Puxin Xu, Zheng Yan, Iliyan Zarov, Yuchen Zhang, Angela Fan, Melanie Kambadur, Sharan Narang, Aurelien Rodriguez, Robert Stojnic, Sergey Edunov, and Thomas
  Scialom.
\newblock Llama 2: Open foundation and fine-tuned chat models, 2023.

\bibitem{frantar2023sparsegpt}
Elias Frantar and Dan Alistarh.
\newblock Sparsegpt: Massive language models can be accurately pruned in one-shot.
\newblock {\em arXiv preprint arXiv:2301.00774}, 2023.

\bibitem{cerebras2023slimpajama}
Daria Soboleva.
\newblock {SlimPajama: A 627B token, cleaned and deduplicated version of RedPajama}, 2023.

\bibitem{kocetkov2022thestack}
Denis Kocetkov, Raymond Li, Loubna Ben~Allal, Jia Li, Chenghao Mou, Carlos Muñoz~Ferrandis, Yacine Jernite, Margaret Mitchell, Sean Hughes, Thomas Wolf, Dzmitry Bahdanau, Leandro von Werra, and Harm de~Vries.
\newblock The stack: 3 tb of permissively licensed source code.
\newblock {\em Preprint}, 2022.

\bibitem{cerebras2024cs3}
James Wang.
\newblock {Cerebras CS-3: the world's fastest and most scalable AI accelerator}, 2024.

\bibitem{deepsparse}
NeuralMagic.
\newblock {DeepSparse}.
\newblock https://github.com/neuralmagic/deepsparse, 2021.

\bibitem{nmvllm}
Robert Shaw and Michael Goin.
\newblock {Bringing the Neural Magic to GPUs}, 2024.

\bibitem{yin2024outlier}
Lu~Yin, You Wu, Zhenyu Zhang, Cheng-Yu Hsieh, Yaqing Wang, Yiling Jia, Mykola Pechenizkiy, Yi~Liang, Zhangyang Wang, and Shiwei Liu.
\newblock Outlier weighed layerwise sparsity (owl): A missing secret sauce for pruning llms to high sparsity, 2024.

\bibitem{jin2022prunings}
Tian Jin, Michael Carbin, Daniel~M. Roy, Jonathan Frankle, and Gintare~Karolina Dziugaite.
\newblock Pruning's effect on generalization through the lens of training and regularization, 2022.

\bibitem{lie2022cerebras}
Sean Lie.
\newblock Cerebras architecture deep dive: First look inside the hw/sw co-design for deep learning: Cerebras systems.
\newblock In {\em 2022 IEEE Hot Chips 34 Symposium (HCS)}, pages 1--34. IEEE Computer Society, 2022.

\bibitem{gsm8k}
Karl Cobbe, Vineet Kosaraju, Mohammad Bavarian, Mark Chen, Heewoo Jun, Lukasz Kaiser, Matthias Plappert, Jerry Tworek, Jacob Hilton, Reiichiro Nakano, et~al.
\newblock Training verifiers to solve math word problems.
\newblock {\em arXiv preprint arXiv:2110.14168}, 2021.

\bibitem{nallapati2016abstractive}
Ramesh Nallapati, Bowen Zhou, Cicero~Nogueira dos santos, Caglar Gulcehre, and Bing Xiang.
\newblock Abstractive text summarization using sequence-to-sequence rnns and beyond, 2016.

\bibitem{ng2015better}
Jun-Ping Ng and Viktoria Abrecht.
\newblock Better summarization evaluation with word embeddings for rouge, 2015.

\bibitem{ding2023enhancing}
Ning Ding, Yulin Chen, Bokai Xu, Yujia Qin, Zhi Zheng, Shengding Hu, Zhiyuan Liu, Maosong Sun, and Bowen Zhou.
\newblock Enhancing chat language models by scaling high-quality instructional conversations, 2023.

\bibitem{tunstall2023zephyr}
Lewis Tunstall, Edward Beeching, Nathan Lambert, Nazneen Rajani, Kashif Rasul, Younes Belkada, Shengyi Huang, Leandro von Werra, Clémentine Fourrier, Nathan Habib, Nathan Sarrazin, Omar Sanseviero, Alexander~M. Rush, and Thomas Wolf.
\newblock Zephyr: Direct distillation of lm alignment, 2023.

\bibitem{hartford2023dolphin}
Eric Hartford.
\newblock {Dolphin}, 2023.

\bibitem{lee2024platypus}
Ariel~N. Lee, Cole~J. Hunter, and Nataniel Ruiz.
\newblock Platypus: Quick, cheap, and powerful refinement of llms, 2024.

\bibitem{luo2023wizardcoder}
Ziyang Luo, Can Xu, Pu~Zhao, Qingfeng Sun, Xiubo Geng, Wenxiang Hu, Chongyang Tao, Jing Ma, Qingwei Lin, and Daxin Jiang.
\newblock Wizardcoder: Empowering code large language models with evol-instruct, 2023.

\bibitem{alpaca_eval}
Xuechen Li, Tianyi Zhang, Yann Dubois, Rohan Taori, Ishaan Gulrajani, Carlos Guestrin, Percy Liang, and Tatsunori~B. Hashimoto.
\newblock Alpacaeval: An automatic evaluator of instruction-following models.
\newblock \url{https://github.com/tatsu-lab/alpaca_eval}, 2023.

\bibitem{open-llm-leaderboard}
Edward Beeching, Clémentine Fourrier, Nathan Habib, Sheon Han, Nathan Lambert, Nazneen Rajani, Omar Sanseviero, Lewis Tunstall, and Thomas Wolf.
\newblock Open llm leaderboard.
\newblock \url{https://huggingface.co/spaces/HuggingFaceH4/open_llm_leaderboard}, 2023.

\bibitem{chen2021codex}
Mark Chen, Jerry Tworek, Heewoo Jun, Qiming Yuan, Henrique~Ponde de~Oliveira~Pinto, Jared Kaplan, Harri Edwards, Yuri Burda, Nicholas Joseph, Greg Brockman, Alex Ray, Raul Puri, Gretchen Krueger, Michael Petrov, Heidy Khlaaf, Girish Sastry, Pamela Mishkin, Brooke Chan, Scott Gray, Nick Ryder, Mikhail Pavlov, Alethea Power, Lukasz Kaiser, Mohammad Bavarian, Clemens Winter, Philippe Tillet, Felipe~Petroski Such, Dave Cummings, Matthias Plappert, Fotios Chantzis, Elizabeth Barnes, Ariel Herbert-Voss, William~Hebgen Guss, Alex Nichol, Alex Paino, Nikolas Tezak, Jie Tang, Igor Babuschkin, Suchir Balaji, Shantanu Jain, William Saunders, Christopher Hesse, Andrew~N. Carr, Jan Leike, Josh Achiam, Vedant Misra, Evan Morikawa, Alec Radford, Matthew Knight, Miles Brundage, Mira Murati, Katie Mayer, Peter Welinder, Bob McGrew, Dario Amodei, Sam McCandlish, Ilya Sutskever, and Wojciech Zaremba.
\newblock Evaluating large language models trained on code.
\newblock 2021.

\bibitem{xiao2023smoothquant}
Guangxuan Xiao, Ji~Lin, Mickael Seznec, Hao Wu, Julien Demouth, and Song Han.
\newblock Smoothquant: Accurate and efficient post-training quantization for large language models.
\newblock In {\em International Conference on Machine Learning}, pages 38087--38099. PMLR, 2023.

\end{thebibliography}

\clearpage

\appendix

\section{Appendix}

\subsection{Sparse Pre-training Ablations}
\label{app:ablations}

\paragraph{Uniform vs. Outlier Weighed Layerwise (OWL) Sparsity Profiles}  Outlier Weighed Layerwise \cite{yin2024outlier} sparsity profile proposed non-uniform layerwise sparsity levels to account for activation outliers in recent, performant LLMs. This allows them to push to non-trivial sparsities (70\%) without re-training at larger model sizes. Following this, we ran an ablation to compare the uniform and OWL profiles for one-shot pruning, followed by re-training. Table \ref{tab:uniform_vs_owl_pretraining} compares the results of  Llama Evaluation metrics, both post-pruning and after retraining for 80B tokens at 40\% pruning. OWL profiles lead to higher recovery post-pruning; however, further re-training degrades the model quality. 

\begin{table}[h]
\caption{Pretrained Llama-2 7B evaluation metrics comparing uniform vs. OWL sparsity profiles at 40\% pruning. Best results are highlighted in bold.}
\label{tab:uniform_vs_owl_pretraining}
\begin{tabular}{|cccccc|}
\hline
\multicolumn{1}{|c|}{\multirow{2}{*}{\textbf{Sparsity Level}}} & \multicolumn{1}{c|}{\multirow{2}{*}{\textbf{Tokens (Type)}}} & \multicolumn{4}{c|}{\textbf{Llama Evaluation Metrics}}                                                                                                                           \\ \cline{3-6} 
\multicolumn{1}{|c|}{}                                         & \multicolumn{1}{c|}{}                                        & \multicolumn{1}{c|}{\textbf{Average}} & \multicolumn{1}{l|}{\textbf{Code}} & \multicolumn{1}{l|}{\textbf{CommonSense Reasoning}} & \multicolumn{1}{l|}{\textbf{World Knowledge}} \\ \hline
\multicolumn{6}{|l|}{\textit{Baseline}}                                                                                                                                                                                                                                                                          \\ \hline
\multicolumn{1}{|c|}{0\%}                                      & \multicolumn{1}{c|}{2T (Training)}                           & \multicolumn{1}{c|}{\textbf{42.1}}    & \multicolumn{1}{c|}{16.4}          & \multicolumn{1}{c|}{63.5}                           & 46.5                                          \\ \hline
\multicolumn{6}{|l|}{\textit{Post-Training Pruning}}                                                                                                                                                                                                                                                             \\ \hline
\multicolumn{1}{|c|}{50\% (Uniform)}                           & \multicolumn{1}{c|}{4M (Calibration)}                        & \multicolumn{1}{c|}{31.1}             & \multicolumn{1}{c|}{2.6}           & \multicolumn{1}{c|}{60.7}                           & 30.1                                          \\ \hline
\multicolumn{1}{|c|}{50\% (OWL)}                               & \multicolumn{1}{c|}{4M (Calibration)}                        & \multicolumn{1}{c|}{\textbf{32.1}}    & \multicolumn{1}{c|}{5.1}           & \multicolumn{1}{c|}{60.7}                           & 30.4                                          \\ \hline
\multicolumn{6}{|l|}{\textit{Sparse Pretraining (Our Approach)}}                                                                                                                                                                                                                                                 \\ \hline
\multicolumn{1}{|c|}{50\% (Uniform)}                           & \multicolumn{1}{c|}{80B (Sparse Training)}                   & \multicolumn{1}{c|}{\textbf{39.8}}    & \multicolumn{1}{c|}{13.9}          & \multicolumn{1}{c|}{62.2}                           & 43.2                                          \\ \hline
\multicolumn{1}{|c|}{50\% (OWL)}                               & \multicolumn{1}{c|}{80B (Sparse Training)}                   & \multicolumn{1}{c|}{39.0}             & \multicolumn{1}{c|}{12.4}          & \multicolumn{1}{c|}{62.2}                           & 42.4                                          \\ \hline
\end{tabular}
\end{table}

\paragraph{Dataset mix} We augmented the existing SlimPajama \cite{cerebras2023slimpajama} dataset with the The Stack \cite{kocetkov2022thestack} dataset, which is a compilation of permissively licensed source code from GitHub. We specifically use the Python split of the data for our sparse pre-training runs from the de-duplicated dataset. We re-balance the data source proportions based on the number of new tokens from the Python split, using the Llama-2 \cite{touvron2023llama} tokenizer. Table \ref{tab:retraining_data_proportions} shows the data source proportions.

\begin{table}[h]
\caption{Data source proportions for SlimPajama and the augmented version with the Python split from The Stack.}
\label{tab:retraining_data_proportions}
\centering
\begin{tabular}{|c|c|c|}
\hline
\textbf{Data Source} & \textbf{SlimPajama} & \textbf{Slimpajama + The Stack (Python)} \\ \hline
ArXiv                & 4.6\%               & 4.2\%                                    \\ \hline
Books                & 4.2\%               & 4\%                                      \\ \hline
C4                   & 26.7\%              & 25.8\%                                   \\ \hline
Commoncrawl          & 52.2\%              & 51.7\%                                   \\ \hline
Github               & 5.2\%               & 4.8\%                                    \\ \hline
StackExchange        & 3.3\%               & 3.1\%                                    \\ \hline
Wikipedia            & 3.8\%               & 3.3\%                                    \\ \hline
The Stack (Python)   & -                   & 3.1\%                                    \\ \hline
\end{tabular}
\end{table}

\end{document}